\documentclass{article}
\usepackage{ijcai20}

\usepackage[hidelinks]{hyperref}
\usepackage[utf8]{inputenc}
\usepackage[small]{caption}

\usepackage{makecell,color,booktabs,amsmath,graphicx,url,soul,times,appendix,dsfont,subfig,algorithm,amsopn}
\usepackage[noend]{algpseudocode}

\usepackage{natbib} 
\usepackage{flushend}

\title{Stereotype-Free Classification of Fictitious Faces}

\author{
Mohammadhossein Toutiaee\and
Soheyla Amirian\and
John A. Miller\And
Sheng Li\\
\affiliations
Department of Computer Science\\
The University of Georgia
\footnote{Contact Author}
\emails
\{hossein, amirian, jamill, sheng.li\}@uga.edu
}

\begin{document}

\maketitle

\begin{abstract}
Equal Opportunity and Fairness are receiving increasing attention in artificial intelligence.
\textit{Stereotyping} is another source of \textit{discrimination}, which yet has been unstudied in literature.
GAN-made faces would be exposed to such discrimination, if they are classified by human perception.
It is possible to eliminate the human impact on fictitious faces classification task by the use of statistical approaches.
We present a novel approach through penalized regression to label stereotype-free GAN-generated synthetic unlabeled images. 
The proposed approach aids labeling new data (fictitious output images) by minimizing a penalized version of the least squares cost function between realistic pictures and target pictures.

\end{abstract}
%
%

\section{Introduction}
\label{sec:intro}

Despite the appealing application and success in Machine Learning tasks, a major field within Artificial Intelligence that began more slower, but has expanded enormously in the recent years is Fairness. 
Discrimination refers to the effect of bias against people's lives due to their membership to different population subgroups. These subgroups are differentiated by the sensitive (protected) attributes recognized by national and international legislation. 
Many applications of machine learning including decision making process can, perhaps unintentionally, result in an unfortunate lack of fairness. As an example their outcomes can asymmetrically deprive (or, enrich) certain subgroups of people with one or more common protected attributes such as race, gender, caste and religion. \cite{joseph2016fairness} enumerated a few examples of applications in policing, hiring and lending where the systematic decision process discrimination might be inevitable. 
Thus, this realization encourages a new era of research in machine learning in the light of fairness affirmative actions.

Scientist have extensively practiced around structured data.
In fairness through unawareness work (FTU) \cite{grgic2016case} proposed a definition of a fair algorithm provided that sensitive attributes $A$ are not explicitly trained in the model.
Experts in individual fairness study (IF) \cite{dwork2012fairness} presented that individuals $i$ and $j$ are similar under a pre-defined metric function if their predictions are similar.
\cite{zemel2013learning} discusses demographic parity/disparity impact (DP) in that $\hat{Y}$ satisfies DP if
\[
P(\hat{Y}|A=0)=P(\hat{Y}|A=1). 
\]
Another group of researchers introduced equality of opportunity (EO) \cite{hardt2016equality} in which they suggest that $\hat{Y}$ satisfies EO if:
\[
P(\hat{Y}=1|A=0,Y=1)=P(\hat{Y}=1|A=1,Y=1),
\]
and the counterfactual fairness study \cite{kusner2017counterfactual} which generalizes the previous work and postulates a mathematical definition, namely as:
\begin{small}
\[
P(\hat{Y}_{A \leftarrow a}=U|A=a,X=x)=P(\hat{Y}_{A \leftarrow a'}=U|A=a,X=x)
\]
\end{small} 
for all $y$ and for any value $a'$ attainable by $A$, the predictor $\hat{y}$ is counterfactually fair. 

Discrimination-aware decision making approaches have been targeted by many researchers, each of which proposing a new quantification of discrimination. 
Thus, fairness definition in predictive models is still controversial with absence of consensus among researchers, and a new school of thought in fairness function is published quite often via research papers to dampen the discrimination effect. 
The variety of approaches leads to a difficulty for evaluation of the progress in the field, and no strengths and weaknesses can be assessed for further recommendations accordingly.

We will focus on generative adversarial network (GAN) in this study for two reasons; first, it is capable of producing fictitious outputs (imaginary images). 
Second, it has inspired a legion of scientists to evolve GAN under the impression of producing more realistic looking data \cite{karras2019style}.
GANs have been widely applied to many domains due to their impressive performance especially on image generation paradigm.
In literature, GANs have only been used to help mitigate bias in data. 
\cite{xu2018fairgan} presented a GAN architecture in which two discriminators and one generator play the adversarial games. 
The generator produces fake data conditioned on the sensitive features, while two discriminators are trained separately to identify whether the generated samples are real or fake, 
and whether or not they're coming from the protected or unprotected group.

We observe that the super-high-quality fictitious images of humans generated by a state-of-the-art GAN, such as Style-GAN \cite{karras2019style}, might be prejudiced by gender or race stereotyping.
In that case, a GAN-made picture of a person with long hair should not be realized necessarily as a woman (gender Stereotyping), and a dark complexion person would not be an African-American individual (skin color Stereotyping).
The protected attributes of a person induce more sensitivity around the subject, and ``Stereotyping'' is another source of discrimination, according to the literature.

The evolutionary of GANs output images (or videos) and Stereotyping issue encourages us to study a new topic in this field.
Although the ``fairGAN'' article \cite{xu2018fairgan} in addition to many other topics in fairness are admiring, no research has previously addressed how to propagate protected attributes (such as gender or race) to GAN-made images without the interference of the human mind judging pictures predominantly based on Stereotypes.

In this work, we propose an interpretable and effective approach to classify synthetic faces of Style-GAN \cite{karras2019style} without symptoms of discrimination.

Typically, fairness scientists have assumed that almost every machine learning algorithm produces outcomes with issues of bias and discrimination.
Thus, they propose a way to manage such issues in the system, either in the pre-processing step or post-processing, based on a pre-defined fairness metric function.
Our method is applied after an image is generated.
For simplicity, we consider two binary sensitive attributes (race and gender), however, our approach can be easily extended to non-binary or more attributes. 

A property of a GAN architecture is the lack of ground truth in the outcomes.
In other words, faces that appear in outputs should not be labeled based on stereotypes such as White woman or Asian man, since these attributes (i.e. race, gender) are self-reported, and in this case, the fictitious images cannot be self-reported by an individual who does not even exist.  
In this paper, we present a novel approach to alleviate the prejudiced view of sensitive attributes (such as gender or race) by minimizing the distance of a given output image from all other realistic input images which contribute in producing that target output.
We call this process ``bring-to-life'' since the attributes of a target image can be described by the most similar real images. 


\section{Preliminary}
\subsection{Background}
Generally, discrimination-free machine learning algorithms are grouped into three broader categories:
\begin{itemize}
    \item The \textit{pre-processing} techniques in which input data is adjusted, so that a target machine learning method deployed upon that data will be fair.
    \cite{feldman2015certifying} proposed a feature modification technique so that each marginal distribution of attribute including sensitive values are all equally likely without touching the training labels.
    \item The algorithm adjustment techniques are those either with improvements to the existing algorithms or designing entirely new algorithms that perform fairly fair under any input.
    This is the main motivation of \cite{zemel2013learning}, in which they combined pre-processing and algorithm adjustment (modification) techniques together to learn a modified version of the data trained through a representation process. 
    \item And any technique that modifies outputs after being processed through a model are grouped as \textit{post-processing}.
    This category has received much attention such as \cite{kamiran2010discrimination} in which, the authors introduce a mechanism for decision tree based methods to modify the labels of leaves after training phase to stabilize the fairness state. 
\end{itemize}

Additionally, there are many metric functions in literature defined for fairness evaluation, and this topic is mainly or partly practiced by scientists.



In this study, we aim to develop a policy through a statistical method to regulate the attributes tagging to generated images without any sign of discrimination against protected groups.
Our method is useful when there is zero information available in the picture for labeling, however, this method is not applicable when the ground truth is available.
Put in another way, our approach is not expected to approve labels that have been already assigned to real images (with ground truth), but we argue that it works well for labeling imaginary pictures (without ground truth).
\begin{figure}[htbp]
    \centering
    \includegraphics[scale=0.5] {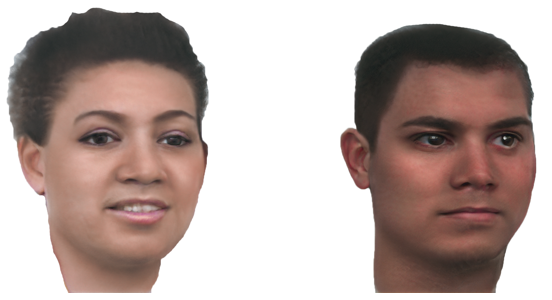}
    \caption{Two examples of fictitious faces generated from adversarial training. Human perception classifies the left and right pictures as White woman and African-American man, respectively. }
    \label{fig:pair}
\end{figure}

\subsection{Related Work}

Scientists are able to generate astonishing pictures of any kind.
The widely used dataset, such as CelebA \cite{liu2015faceattributes}, is utilized by many GAN practitioners to create super natural imaginary pictures with 40 face attributes. 
One can argue that how can a face be considered to be a ``woman'' or ``white'' when the images are not existing in life? 
When output images are produced in a GAN setting, no ground truths are available for the labels of generated images. 
This constraint is due to the fact that GANs try to clone the latent distributions of inputs with different realizations.
In other words, the outputs of a GAN are generated from approximately the same distribution of the inputs but in different samples.

On the other hand, \textit{sex} and \textit{gender}, for instance, are used to refer to biological distinctions between males and females, and psychological characteristics that are learned through the socialization process, respectively \cite{o1997gender}. 
Furthermore, \textit{stereotypes} and discrimination is a well-studied research in  Law and Social Studies where the central arguments are that gender or race stereotypes, for example, lead to biased decisions, discrimination and misrepresentations \cite{heilman2012gender}, (see Figure \ref{fig:pair}).
Therefore, any self-report attributes such as race or gender are not inferred through descriptive features that are recognized from a face in a given GAN-made picture. 
There is no relation between an individuals portrayal aspects and their gender or race. As an example, it cannot be implied that an individual with make-up and long hair is a female, unless it is report by themselves.

Thus, this study attempts to address this non-trivial problem in the next sections.

\subsection{Adversarial Network Architecture}
Generative Adversarial Networks \cite{goodfellow2014generative} are machine learning models that can imagine new samples. 
A generative model $\mathcal{G}$ trained on training data $\mathcal{X}$ sampled from some true distribution $\mathcal{D}$ is one which, given some standard random distribution $\mathcal{Z}$, produces a distribution $\mathcal{D}'$ which is close to $\mathcal{D}$, according to a pre-defined metric function.

\begin{figure}[htbp]
    \centering
    \includegraphics[width=\linewidth] {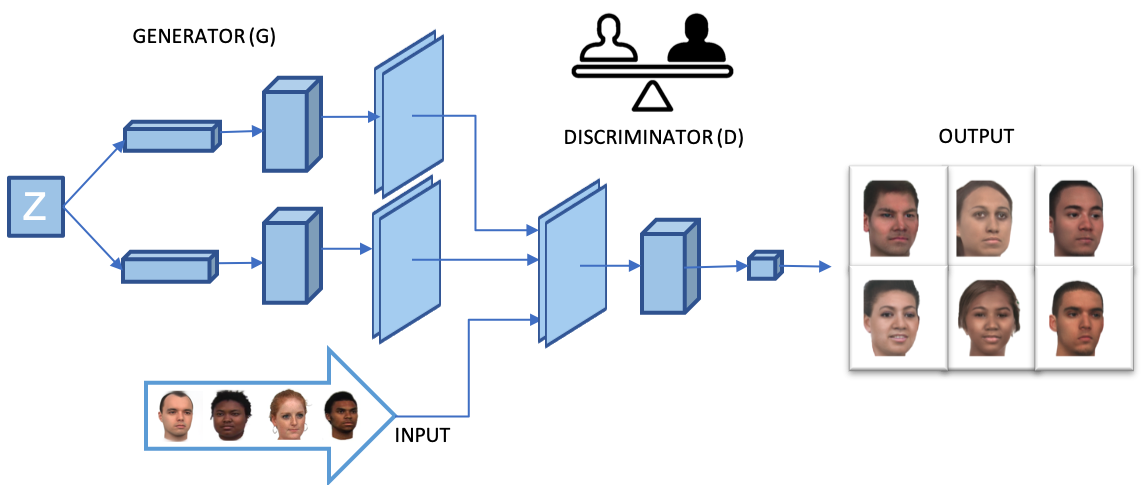}
    \caption{An adversarial training architecture for generating imaginary images.}
    \label{fig:GANArch}
\end{figure}

The objective function of $D$ and $G$ are respectively defined as:
\[
\max_D \quad \mathds{E}_{x\sim P_{data}} [logD(x)]+\mathds{E}_{z\sim P_{z}}[log(1-D(G(z)))], 
\]

\[
\min_G \quad \mathds{E}_{z\sim P_{z}} [log(1-D(G(z)))].
\]
During the training process, the discriminator is shown real images from the training set \%50 of the time, and fake images from the generator the other \%50 of the time.
Over time, the generator is forced to produce more realistic outputs in order to fool the discriminator.
The generator takes random noise values $z$ from a prior distribution $P_z$ and maps them to output values $x$ via function $G(z)$. 
Wherever the generator maps more values of $z$, the probability distribution over $x$, represented by the model, becomes denser.
The discriminator outputs high values wherever the density of real data is greater than the density of generated data.
Thus, the GAN (Figure \ref{fig:GANArch}) is formulated as $\min_G \max_D V(G,D)$, namely as:
\[
V(G,D) = \mathds{E}_{x\sim P_{data}} [logD(x)]+\mathds{E}_{z\sim P_{z}}[log(1-D(G(z)))].
\]
There are many variants on GAN extended by machine learning enthusiasts, while some of them are very prominent. 
We chose Style-GAN as the main architecture to bear imaginary faces, then we classify faces (bring them to life).

\noindent\textbf{Style-GAN} \cite{karras2019style}. 
The researchers in Style-GAN tried to improve the quality of output images by proposing an alternative generator architecture in adversarial training.
They claimed that their new generator is superior to the traditional GAN architecture.
We utilize Style-GAN in this study, since it produces high quality of images.

\section{Proposed Approach}

The purpose of imaginary face classification is to copy the sensitive attributes of real faces to unreal faces in such a way that this process is unbiased. 
Linear regression and unbiasedness in our work are very similar, in that both try to minimize the average distances of data points to a particular line, which is the regression line (or fair line).
The position of this line leads us to classify unreal images fairly.
In our study, Ridge regression would help locate this fair line.
The Ridge regression model we construct (Figure \ref{fig:emd}) would determine the attributes of GAN-made faces by finding the relation between real and imaginary faces.
Coefficients estimation, as a process of minimizing the cost function, would ultimately aid to perform the task. 
More details are provided in the next sections.

\subsection{Stereotype-Free Classification Method}
Each coefficient $X_j$ represents $j$-th GAN-made image that is supposed to be labeled.
Thus, a dataset containing a number of observations $n_i$ (real images) with several features $X_j$ (imaginary images), and a response variable $y_i$ (a protective attribute) which is given as the ground truth per observation form a classic dataset which all the statistics assessments and principles can be applied to.
The Earth Mover Distance (EMD) which reflects the similarity between content-base images, can be computed by various effective algorithms in image retrieval domain.
EMD helps us capture the relation between $X_j$ and $y_i$, and it is discussed in the following.



\begin{algorithm}[ht]
\caption{Stereotype-free Classification Method}\label{mrlsh}
\textbf{INPUT:} $n_i$, $X_j$ and $y_i$\\
\textbf{OUTPUT:} Sign of $\beta_j$ and Unreal Face Classification
\begin{algorithmic}[1]
\ClassInfo{MyProcedure}{}
\State \textbf{EMD Extraction}
\State \textbf{method} \textit{EMD}$(n_i,X_j)$
\State \hspace{1 cm} \textbf{for} $i,j=1$ \textbf{To} $I,J$ 
\State Compute similarity between $n_i$ and $X_j$
\State \hspace{1 cm} \textbf{end for}
\State \textbf{Ridge Regression Estimate}
\State
    $
    \underset{\beta}{\mathrm{argmin}} \sum_{i=1}^n (Y_i - \beta_0 - \sum_{j=1}^k \beta_j X_{ij})^2 + \lambda \sum_{j=1}^k \beta^2_j
    $
\State \hspace{1 cm} \textbf{Goal:} $\hat{\beta}_j$ Estimate
\State \textbf{Classification Task}
\State \hspace{1 cm} Classify Unreal Faces by Sign of $\hat{\beta}_j$:
\State \hspace{1 cm} (-) sign means dissimilar to $y_i$ class reference
\State \hspace{1 cm} (+) sign means similar to $y_i$ class reference
\end{algorithmic}
\end{algorithm}

\subsection{Similarity Measure}
Our proposed method refers to the classification of unreal faces based on the similarity between real and unreal faces.
We evaluate image similarity between $X_j$ and $n_i$ by \textit{Earth Mover’s Distance} (EMD), which has been studied in computer vision and image retrieval for a long time \cite{rubner2000earth}.
Discrete Kantorovich formulation (i.e. EMD), which arises from the idea of optimal transport, provides better distinction between the images approximated by the histograms, as opposed to other conventional measures such as Euclidean distance.

Formally, the EMD between two histogram images $q = (q_1,\ldots,q_n)$ and $p = (p_1, \ldots, p_n)$ is defined as follows:

\begin{eqnarray*}
\text{EMD}(q,p) &=& \min_F \sum_{i=1}^{n}\sum_{j=1}^{n}f_{i,j}c_{i,j},\\
\text{such that}\quad \forall i,j \in [1,n]&:& f_{i,j} \geq 0,\\
\forall i \in [1,n]&:& \sum_{j=1}^n f_{i,j}=q_i,\\
\forall j \in [1,n]&:& \sum_{i=1}^n f_{i,j}=p_j
\end{eqnarray*}
where $q$ and $p$ are assumed to be normalized such that $\sum_{i=1}^n q_i=\sum_{i=1}^n p_i$, and $F$ is a \textit{flow matrix}, where $f_{i,j}$ indicates flow (i.e. earth) to move from $q_i$ to $p_j$, and a \textit{cost matrix} $C$, where $c_{i,j}$ models cost of transporting flow from $i-$th bin to the $j-$th bin.
EMD$(q,p)$ is the minimum cost needed to move $q$ to $p$, and EMD$(q,p)$ is equal to EMD$(p,q)$ when the cost matrix $C$ is symmetric.

\begin{figure}[htbp]
    \centering
    \includegraphics[width=\linewidth]
    {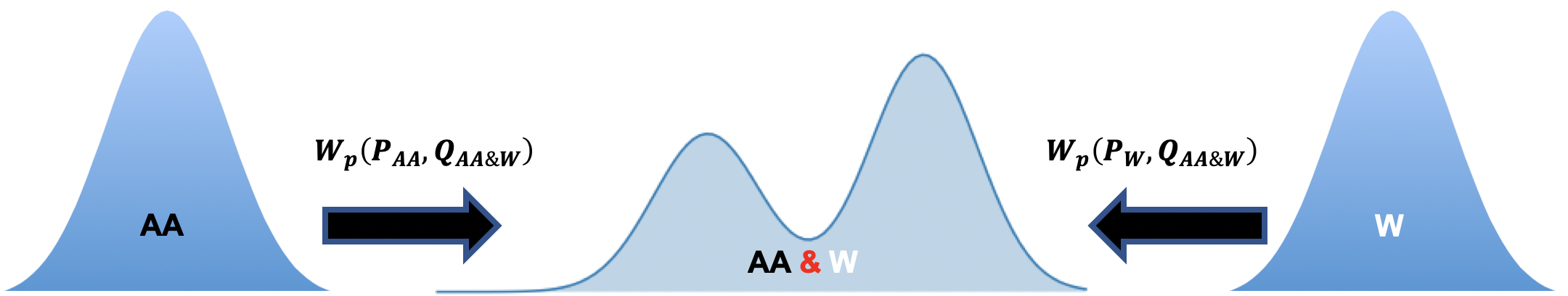}
    \caption{Earth Mover Distance (or 1D-Wasserstein) as similarity metric between images. ``W'' and ``AA'' are representing White and African-American, respectively.}
    \label{fig:emd}
\end{figure}

This similarity function (EMD) is mainly used as the metric comparison between imaginary output images and actual input images in this study (see Figure \ref{fig:emd}). 
This source of variability is obtained per  artificial output image, and it is fed into a \textit{Ridge Regression} model for detecting the most significant variations. 

\subsection{Implementation Method}
\subsubsection{Logistic Function}

The ordinary logistic regression with binary response is given by the probability of the response success:

\[
P(y_i = 1) = \pi_i = \frac{e^{x_i\beta}}{1+e^{x_i\beta}}
\]
where $x_i$ is the $i-$th row of a matrix of $n$ observations, and $\beta$ is the column vector of the regression coefficients. 
The parameters are estimated by maximizing the log-likelihood function:


\begin{eqnarray*}
l(\beta) &=& \sum_{i=1}^n [y_ix_i\beta - log(1+e^{x_i\beta})]
\end{eqnarray*}

One can consider protected attributes to be modeled as binary response where $\pi$ is the probability of success for the corresponding attribute (e.g. race, gender).
We use this paradigm for ethnicity (African-American vs. White) and gender (Male vs. Female) separately, as the binary responses in this study (Figure \ref{fig:pair}).
\subsubsection{Ridge Estimation}

In Ridge regression, one finds the set of $\beta_j'$s that minimize the expression:

\[
\sum_{i=1}^n (Y_i - \beta_0 - \sum_{j=1}^k \beta_j X_{ij})^2 + \lambda \sum_{j=1}^k \beta^2_j
\]
where $\lambda$ is known as ``tuning parameter''. 
If $\lambda = 0$, this simply reproduces the least-squares estimator for the full ($k-$variable) model. 
As $\lambda$ becomes large, the $\beta_j'$s (other than $\beta_0 = \bar{Y}$) collapse to zero, so that exactly the null model emerges.

The idea is to choose $\lambda$ so as to keep important variables' $\beta_j$'s at high magnitude and to shrink the others to near zero. 
The name ``Ridge Regression'', when the idea was originally proposed by Hoerl, arose from the fact that this method frequently helped in cases where there was strong multi-collinearity between two or more predictor variables ($X_j$). In such cases, it is hard to find a true maximum to log-Likelihood function (or equivalently, a true minimum to the negative log-likelihood function) because the max/min lies along a ridge/valley. Incorporating the penalty term allows a true peak or minimum to appear.


\begin{table*}[htbp]
\begin{minipage}{0.5\textwidth}
\centering
\begin{tabular}{|l|c|c|}
 \hline
 \textbf{Method(M)(F)(P-value)} & \textbf{\#detection} & \textbf{Biased?} \\
 \hline
  \text{Human-Perception(31)(33)(0.9)} & \text{\thead{64}} & \text{\thead{No}} \\
 \hline
 \text{Face Classification(31)(33)(0.9)} & \text{\thead{64}} & \text{\thead{No}} \\
 \hline
  \text{Ridge(34)(30)(0.7)} & \text{\thead{64}} & \text{\thead{\textcolor{green}{No}}} \\
 \hline
 \end{tabular}
\end{minipage}
\begin{minipage}{0.5\textwidth}

\centering
\begin{tabular}{|l|c|c|}
 \hline
 \textbf{Method(AA)(W)(P-value)} & \textbf{\#detection} & \textbf{Biased?} \\
 \hline
  \text{Human-Perception(18)(46)(0.0006)} & \text{\thead{64}} & \text{\thead{\textcolor{red}{Yes}}} \\
 \hline
 \text{Face Classification(22)(36)(0.043)} & \text{\thead{58}} & \text{\thead{\textcolor{red}{Yes}}} \\
 \hline
  \text{Ridge(36)(28)(0.38)} & \text{\thead{64}} & \text{\thead{\textcolor{green}{No}}} \\
 \hline
 \end{tabular}
\end{minipage}
\caption{[Left] Gender: Male (M) Vs. Female (F). [Right] Race: African-American (AA) Vs. White (W) Results obtained from several evaluated methods for Gender (left) and Race (right). The first and second number per method indicate the number of instances that method has detected. The third number (P-value) shows the probability of the null hypothesis being true under the statistical threshold (0.05) to test whether the corresponding method prefers one particular attribute over another. If the related P-value exceeds the threshold (0.05), then one concludes that the impartial preference is rejected and the corresponding method is biased against the minority. As both tables confirm, Ridge method tends to propagate labels to all images (64 out of 64) without any sign of discrimination (p-value $>$ 0.05).}
\label{tab:race}
\end{table*}


\subsubsection{Ridge Logistic Regression}

The Logistic Ridge Regression estimator depends on the choice of a tuning parameter $\lambda \geq 0$, and the coefficients parameters
are obtained when the following slightly different log-likelihood function with extra $L_2$ Ridge penalty is maximized. 
The constrained maximization equation is as follows:

\[
l_{\lambda}^R (\beta) = \sum_{i=1}^n [y_ix_i\beta - log(1+e^{x_i\beta})] - \lambda \sum_{j=1}^p \beta^2_j
\]
where $\beta_j$ is the unlabeled output coefficient, $x_i$ is the EMD value, $y_i$ is binary response, $n$ is the number of labeled inputs, $p$ is the number of unlabeled outputs, and $\lambda,\alpha$ are the hyper-parameters obtained through Cross-Validation process.

\subsection{Implementation Process}

The backbone of our approach is based on image similarity calculated between the imaginary outputs and every real input image.
One positive side-effect of the similarity distance between inputs and outputs is a tendency of our approach to be less prone to classify unfairly.
The earth mover distance (EMD) discussed earlier provides a similarity measure between images.
EMD is chosen as a similarity metric, since it matches better with the human perception of differences compared to other distances such as Euclidean distance or $\chi^2$ divergence \cite{rubner2001empirical}.
The EMD level obtained per output image encodes the relation between the unlabeled imaginary image and previously labeled images fed into the Adversarial training network (Figure \ref{fig:GANArch}).
\begin{figure}[htbp]
    \centering
    \includegraphics[width=\linewidth] {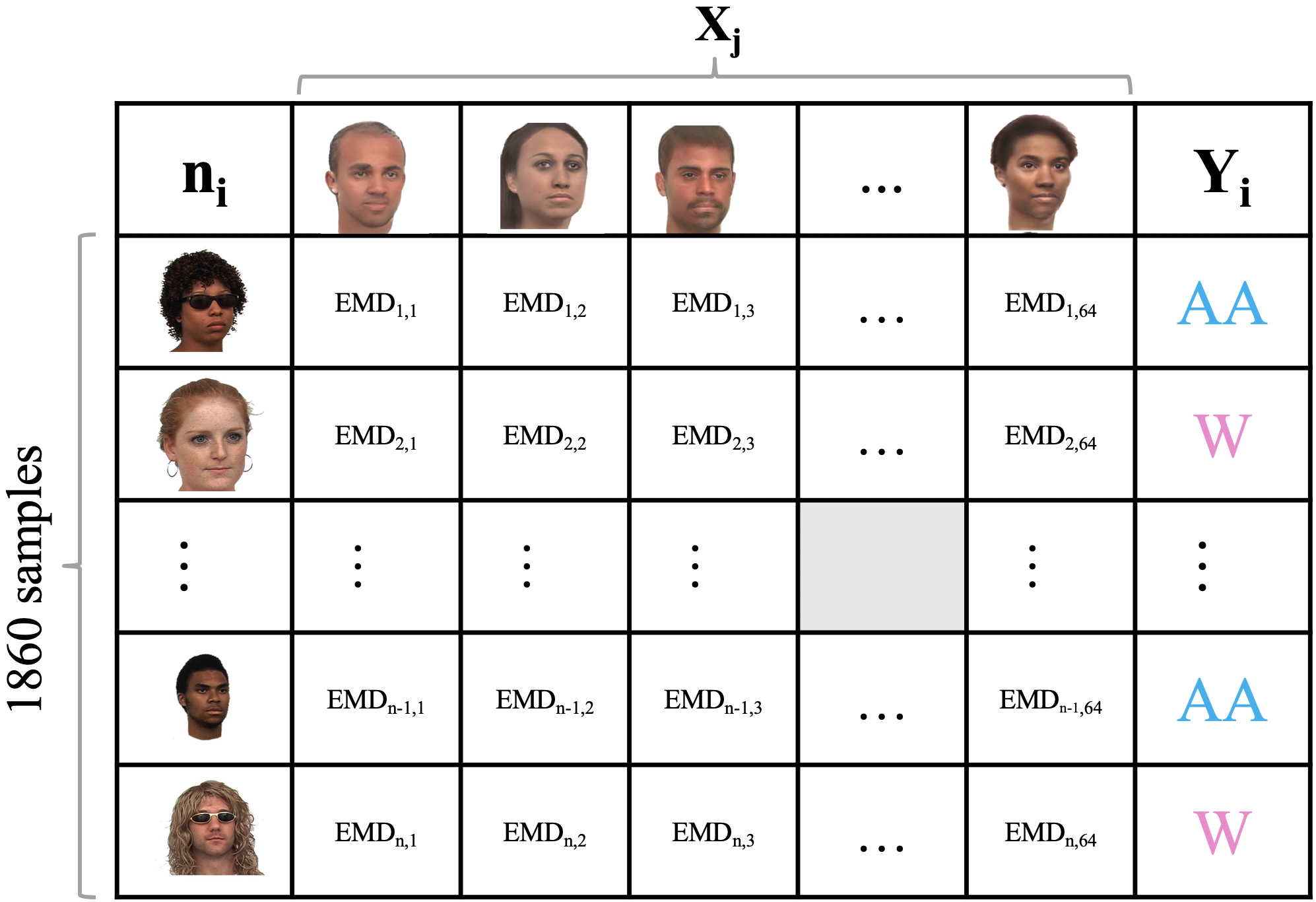}
    \caption{Fictitious face classification process. $y_i$ is true labels, $X_j$ is fictitious faces and EMD refers to Earth Mover Distance (similarity measure).}
    \label{fig:EMD1}
\end{figure}
Having extracted all the EMD values, one can interpret the problem as a classification task, in which $n$ is the number of observations (i.e. labeled inputs), $X$ is the number of features (i.e. unlabeled outputs) and $y$ is the binary response (i.e. gender or race).
As the number of unlabeled images increase, the dataset may suffer from the curse of dimensionality, and all the principles affected in high-dimensional dataset are enforced.
This encourages us to choose a model which is appropriate for a high-dimensional classification (labeling) problem such as penalized regression family discussed in the previous sections. We primarily adopt the Ridge Logistic Regression model with extra regularization term (penalty) to handle such high-dimensional data. One positive property of Ridge Logistic Regression is that it preserves the features in the model which ensures all the imaginary faces are tagged by the true attributes.

\noindent\textbf{Fairness Evaluation:} Race preference study of GANs is another contribution of this work as it measures which race is preferred over another. 
In comparison with the baseline and face classification methods, we construct the null hypothesis $H_0$ to test statistically whether the protected attributes are equally preferred vs. the alternative ($H_a$) that they are not equally preferred across all the methods (Table \ref{tab:race}).
Then a sign test, with an $\alpha$-level of 0.05, of the null hypothesis ($H_0$) is evaluated statistically by the two-tailed P-values in each approach, calculated by Binomial distribution.

\begin{figure}[!ht]
    \centering
    \includegraphics[width=\linewidth] {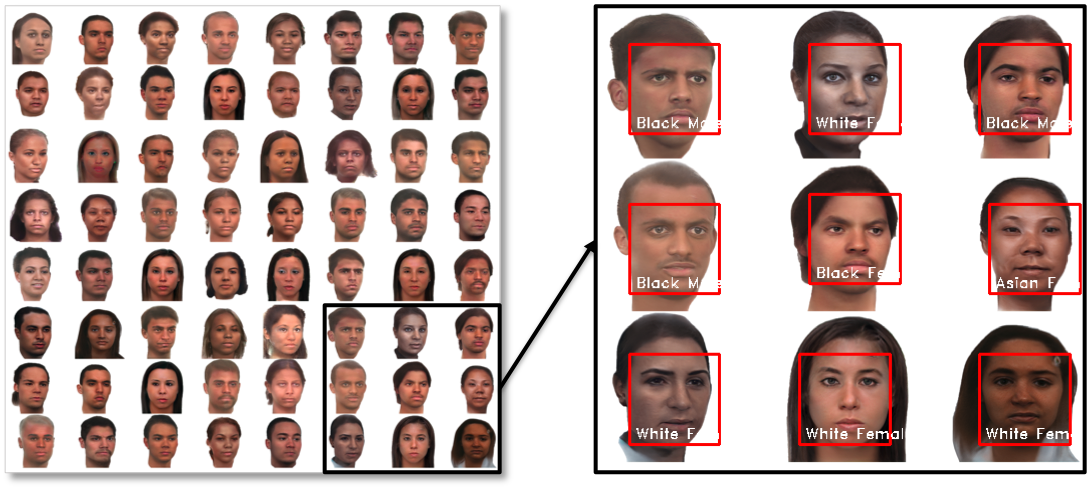}
    \caption{Left: Imaginary faces generated from GAN. Right: Attribute tagging by face classification.}
    \label{fig:raceDetect}

\minipage{0.5\linewidth}
  \captionsetup{width=0.9\linewidth}
  \includegraphics[width=\textwidth]{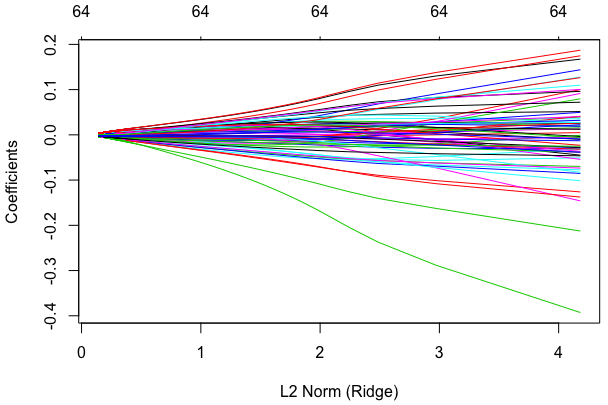}
  \caption*{\scriptsize{(a)}}
\endminipage\hfill
\minipage{0.5\linewidth}
  \captionsetup{width=0.9\linewidth}
  \includegraphics[width=\linewidth]{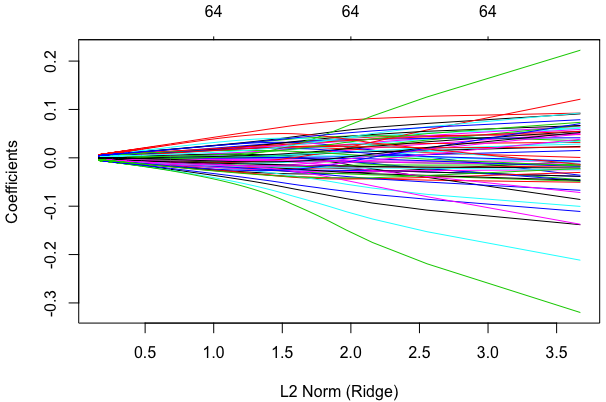}
  \caption*{\scriptsize{(b)}}
\endminipage\hfill

\minipage{0.5\linewidth}
  \captionsetup{width=0.9\linewidth}
  \includegraphics[width=\textwidth]{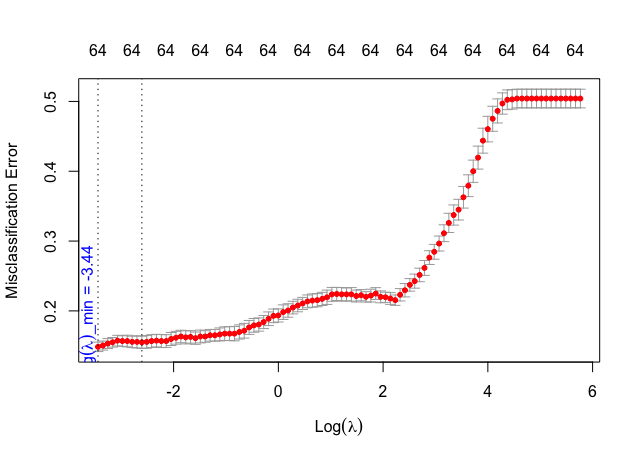}
  \caption*{\scriptsize{(c)}}\label{fig:landaG}
\endminipage\hfill
\minipage{0.5\linewidth}
  \captionsetup{width=0.7\linewidth}
  \includegraphics[width=\linewidth]{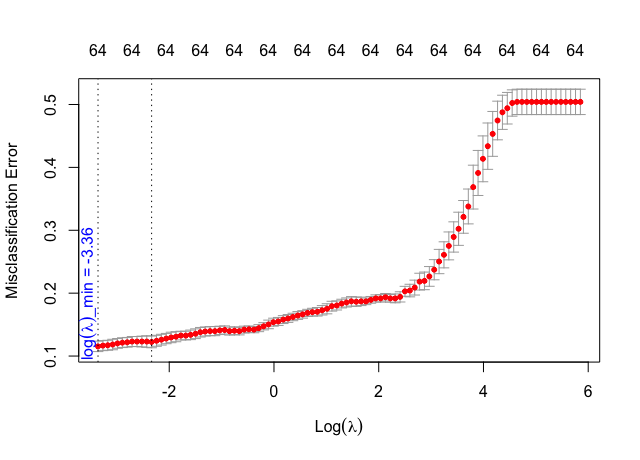}
  \caption*{\scriptsize{(d)}}\label{fig:landaR}
\endminipage\hfill
\caption{\scriptsize{(a) Coefficients estimate for gender in Ridge regression. (b) Coefficients estimate for race in Ridge regression. (c) Lambda estimate for gender in Ridge regression. (d) Lambda estimate for race in Ridge regression.}}
\label{fig:coeffR}
\end{figure}

\begin{figure}[!ht]
\minipage{0.5\linewidth}
  \captionsetup{width=0.85\linewidth}
  \includegraphics[width=\textwidth]{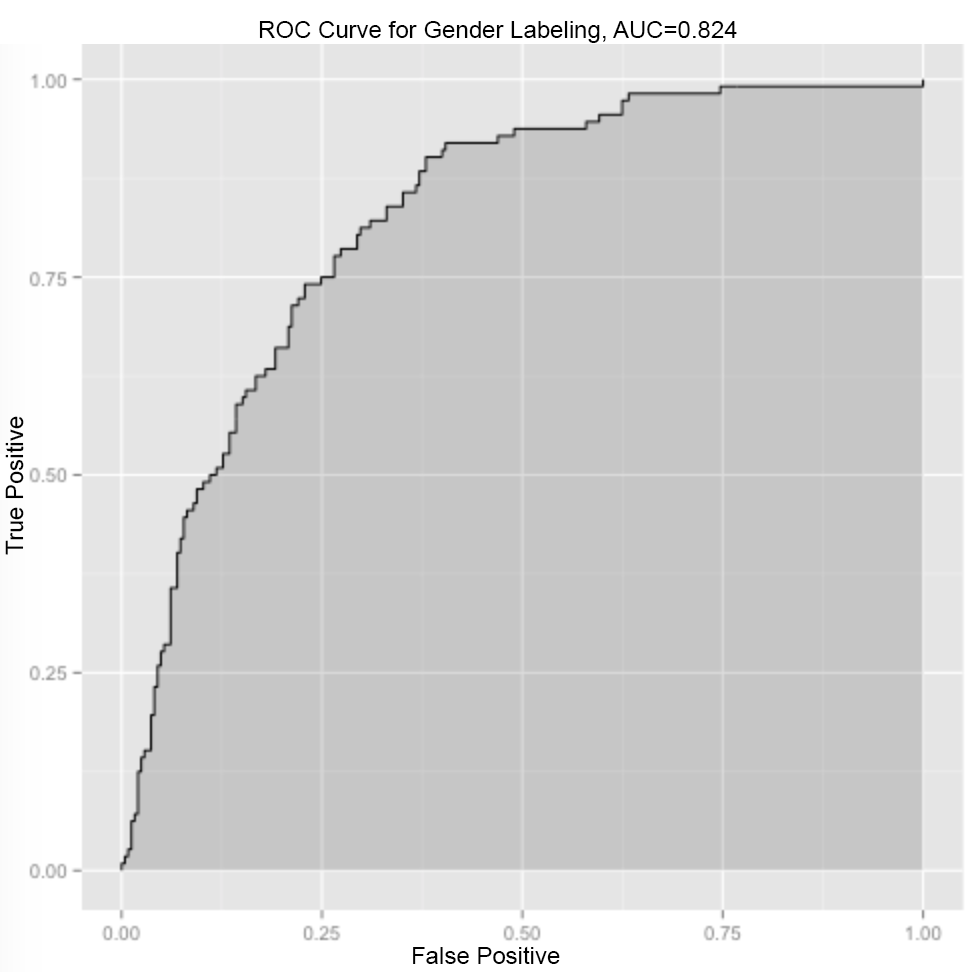}
  \caption*{\scriptsize{(a)}}
\endminipage\hfill
\minipage{0.5\linewidth}
  \captionsetup{width=0.85\linewidth}
  \includegraphics[width=\linewidth]{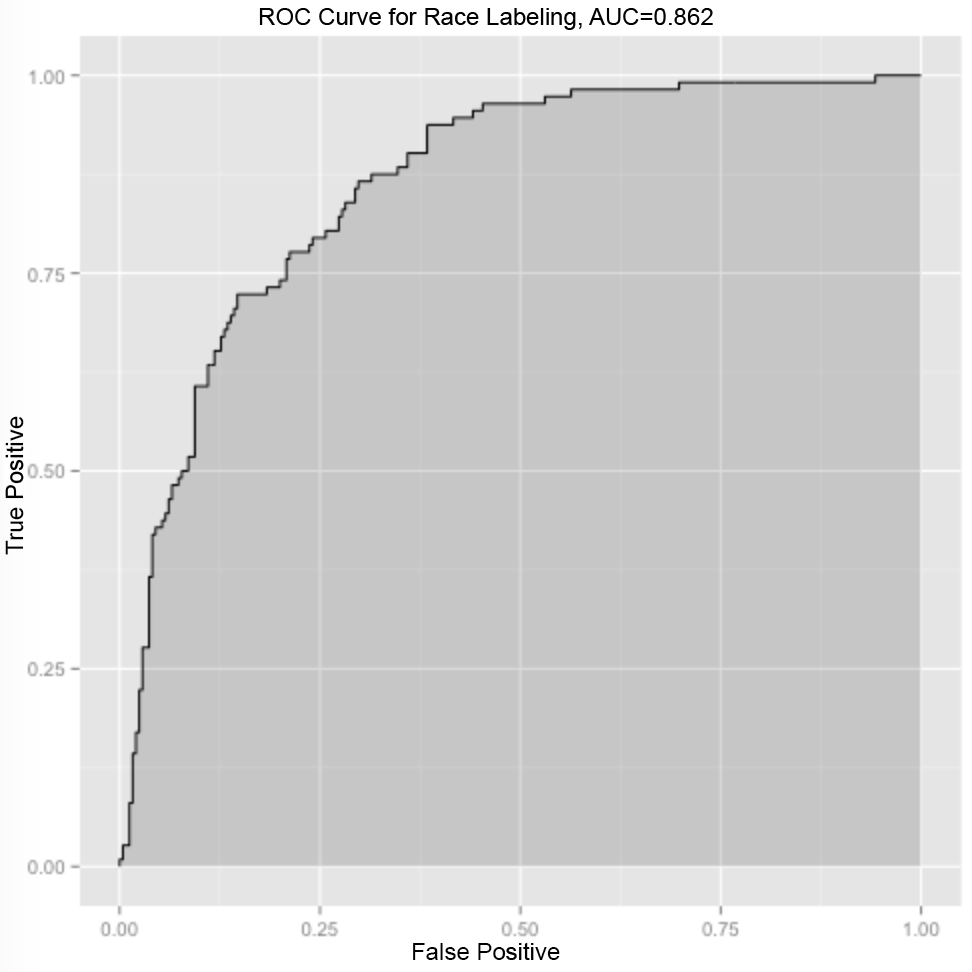}
  \caption*{\scriptsize{(b)}}
\endminipage\hfill
\caption{(a) ROC curve for gender labeling in Ridge regression. (b) ROC curve for race labeling in Ridge regression. The AUC scores greater than \%82 obtained for both tasks (i.e. gender and race labeling) from Ridge regression supports this idea that the proposed technique for classifying faces is reliable.}
\label{fig:rocR}
\end{figure}

\section{Experiments}
\noindent\textbf{Baselines}. To evaluate the effectiveness of our method, we compare its performance with human perception (by 10 different individuals) and pre-trained face classification\footnote{https://github.com/wondonghyeon/face-classification}. 

\noindent\textbf{Dataset:} The dataset Face Place\footnote{http://wiki.cnbc.cmu.edu/Face\_Place} contains 930 images of African-American and Caucasian each with a resolution of $128\times 128$. 
Two goals are ultimately pursued to determine first, the imaginary faces created by GAN are labeled fairly Man or Woman, African-American or White, and second, to compare it with the baseline approaches as to whether the two races or genders are equally preferred or not.
The synthetic images generated from STYLE-GAN are given in Figure \ref{fig:raceDetect}.
The results presented in the figure reveal that our produced faces are outstandingly looking natural to viewers.
One reason is that STYLE-GAN utilizes more sophisticated architecture with better representation training.
So it can achieve state-of-the-art performance.
\subsection{Model Performance}
For ease of exposition, we will focus on penalized regression family when the response variable is binary (i.e. race with two classes), but our method can be adopted for other types of classifier models as well.
In this experiment, we evaluate the effectiveness of the proposed methods in terms of classification accuracy and efficiency and compare all the obtained results together. 
The results are summarized in Table \ref{tab:race} and Figure \ref{fig:raceDetect}.

Tables \ref{tab:race} indicates, human perception as a baseline approach is biased in favour of White group (p-value $<$ 0.05 indicating that $H_0$ is rejected).
This is the result of the biased human mind in determining the race of the artificial faces.

Although the face classification method performs on par with human perception in the gender attribute, it raises this concern that it has been trained unfairly against minorities. 
In our experiments, our approach (Ridge) tends to label all the imaginary faces (64), without any signs of bias presented in Figure \ref{fig:coeffR} and \ref{fig:rocR}.

\subsection{Experimental Results}

According to the proposed method, White and African-American people, Man and Woman are labeled based on negative and positive signs obtained through a weights estimation process.
The interpretation of the sign is entirely based on how the response variable ($y$) is referenced in the model.
In the race labeling process, for example, $y$ accepts values of 0 and 1 for African-American and Caucasian, respectively.
So the negative signs that appear in the coefficients estimation represent African-American, because negative relationship in the equation determines more similarity with the referenced response variable.
We are interested in studying as to whether the null hypothesis ($H_0:$ the two races are equally preferred) evaluated by P-value at $\alpha$-level of 0.05 is rejected or not.
Same $H_0$ can also be evaluated for the gender attribute.
The two-tailed p-values in each case can be found from the same Binomial distribution discussed in the previous sections.
The calculated p-values would determine whether the $H_0$ hypotheses is rejected, which indicates that the corresponding method violates fairness policy.

In our experiments, Ridge Logistic Regression tends to yield stereotyping-free labeling in generated unlabeled images in both tasks, without any signs of performance degradation.
These evidences are provided in Table \ref{tab:race}.

\section{Conclusion and Future Work}

There is no ground truth for outputs of GAN-made images of fictitious people, since GANs generate imaginary data.
Images generated from GANs are mostly classified by human perception, and human perception suffers from stereotyping paradigm.
We discussed that stereotype is another source of discrimination, which leads to behavioural bias against sub-groups of people. 
We presented a stereotype-free labeling approach to eliminate such discrimination as a result of human perception.
This is a new angel of view which stimulates a viewer's attention by looking at the super natural imaginary faces.
Our method is useful when there is no ground truth for faces, and it is not applicable on images with true labels.
We also utilized the Earth Mover Distance (EMD) as the similarity metric to evaluate the classification process, and the artificial output images are ultimately labeled according to the relations between predictor variables (fictitious faces) and the response variable(s) (real faces attribute).
The results revealed that Ridge Logistic Regression labeled fairly all the imaginary images due to its shrinkage property, while human perception and the deep trained face classification are biased in favour of one sub-group.
Based on our experiment, it is becoming clear that human perception is not a reliable source for judging synthetic faces, and it is governed by stereotypes.
In general, we expect that this study for stereotype-free labeling of the GAN-made faces will provide interesting avenues for future work.
We plan to expand this research into other applications and domains by use of other unbiased techniques.
\bibliographystyle{plain}
\bibliography{ijcai20}





\end{document}